\documentclass[sigconf]{acmart}

\AtBeginDocument{%
  \providecommand\BibTeX{{%
    \normalfont B\kern-0.5em{\scshape i\kern-0.25em b}\kern-0.8em\TeX}}}

\setcopyright{acmcopyright}
\copyrightyear{2018}
\acmYear{2018}
\acmDOI{10.1145/1122445.1122456}

\acmConference[Woodstock '18]{Woodstock '18: ACM Symposium on Neural
  Gaze Detection}{June 03--05, 2018}{Woodstock, NY}
\acmBooktitle{Woodstock '18: ACM Symposium on Neural Gaze Detection,
  June 03--05, 2018, Woodstock, NY}
\acmPrice{15.00}
\acmISBN{978-1-4503-XXXX-X/18/06}

\usepackage{array}
\usepackage{amsmath}
\usepackage{amsthm}
\usepackage{mathrsfs}
\usepackage{bm}
\usepackage{booktabs}
\usepackage{multirow}
\usepackage{caption}
\usepackage{subfigure}
\usepackage{graphicx}
\usepackage{color}
\usepackage{hyperref}
\usepackage{algorithm}
\usepackage{algorithmic}

\usepackage{wrapfig}

\hypersetup{hidelinks, colorlinks=true, allcolors=black, pdfstartview=Fit, breaklinks=true}

\usepackage{algorithm,algorithmic}
\renewcommand{\algorithmiccomment}[1]{\bgroup\hfill\footnotesize//~#1\egroup}

\newcommand{\problemname}{\mbox{RMLE}}
\newcommand{\methodname}{\mbox{ICPA}}

\def\calL{\mathcal{L}}

\def\calP{\mathcal{P}}

\def\calR{\mathcal{R}}

\def\calD{\mathcal{D}}

\def\calG{\mathcal{G}}
\def\calA{\mathcal{A}}

\def\a{\mathbf{a}}
\def\b{\mathbf{b}}

\def\x{\mathbf{x}}

\def\y{\mathbf{y}}
\def\z{\mbox{\boldmath$z$}}

\def\w{\mathbf{w}}

\def\v{\mathbf{v}}

\def\t{\mbox{\boldmath$t$}}

\def\nnu{\mbox{\boldmath$\nu$}}
\def\llambda{\mbox{\boldmath$\lambda$}}

\def\ttheta{\mbox{\boldmath$\theta$}}

\def\xxi{\mbox{\boldmath$\xi$}}

\def\calLL{\mbox{\boldmath$\mathcal{L}$}}
\def\ppi{\mbox{\boldmath$\pi$}}

\newtheorem{theorem}{Theorem}

\newtheorem{definition}{Definition}

\definecolor{Gray}{gray}{0.85}
\definecolor{LightCyan}{rgb}{0.88,1,1}

\begin{document}

\title{Incentive Compatible Pareto Alignment for 
Multi-Source Large Graphs}

\author{Jian Liang$^1$, Fangrui	Lv$^2$, Di Liu$^1$, Zehui Dai$^1$, Xu Tian$^1$, Shuang Li$^2$, Fei Wang$^3$, Han Li$^1$}
\affiliation{%
  \institution{$^1$Alibaba Group, China\\$^2$Beijing Institute of Technology, China\\$^3$Department of Population Health Sciences, Weill Cornell Medicine, USA}
\country{}
}
\email{{xuelang.lj, wendi.ld, zehui.dzh, xu.tian, lihan.lh}@alibaba-inc.com}
\email{{fangruilv, shuangli}@bit.edu.cn}
\email{few2001@med.cornell.edu}




\begin{abstract}
In this paper, we focus on learning effective entity matching models over multi-source large-scale data. For real applications, we relax typical assumptions that data distributions/spaces, or entity identities are shared between sources, and propose a Relaxed Multi-source Large-scale Entity-matching (\problemname) problem.
Challenges of the problem include 1) how to align large-scale entities between sources to share information and 2) how to mitigate negative transfer from joint learning multi-source data. What's worse, one practical issue is the entanglement between both challenges. Specifically, incorrect alignments may increase negative transfer; while mitigating negative transfer for one source may result in poorly learned representations for other sources and then decrease alignment accuracy. 
To handle the entangled challenges, we point out that 
the key is to optimize information sharing first based on Pareto front optimization, by showing that information sharing significantly influences the Pareto front which depicts lower bounds of negative transfer.
Consequently, we proposed an Incentive Compatible Pareto Alignment (\methodname) method to first optimize cross-source alignments based on Pareto front optimization, then mitigate negative transfer constrained on the optimized alignments.
This mechanism renders each source can learn based on its true preference without worrying about deteriorating representations of other sources. Specifically, the Pareto front optimization encourages minimizing lower bounds of negative transfer, which optimizes whether and which to align. In detail, we adopt graph neural networks to handle data sparsity in each source and a scalable alignment based on sliced graph matching. 
Comprehensive empirical evaluation results on four large-scale datasets are provided to demonstrate the effectiveness and superiority of ICPA. Online A/B test results at a search advertising platform also demonstrate the effectiveness of ICPA in production environments. We also release an International Entity Graph (IEG)\footnote{\url{https://tianchi.aliyun.com/dataset/dataDetail?dataId=89912}} dataset to facilitate future research. 
\end{abstract}

\ccsdesc[500]{Computing methodologies~Multi-task learning}

\keywords{Multi-source Large Graphs, Transfer Learning, Negative Transfer}
\maketitle


\section{Introduction}\label{sec:intro}
Entity matching is the task of confirming the correlation between two entities based on certain correlation criteria. Large-scale entity matching is common problem in real-applications, including identity recognition~\cite{zhao2003face,bedagkar2014survey},
information retrieval~\cite{baeza1999modern}, and recommendation systems~\cite{bobadilla2013recommender}. However, the commonly-encountered data-sparsity problem often hurts the 
generalization of the learned matching model~\cite{grvcar2005data,pan2010transfer}, which is usually due to lacking annotation of matching relationships. To tackle this problem, Multi-source Entity-Matching (MEM) which exploits data from one or several auxiliary sources acts as a mainstream solution~\cite{chen2020multi,lu2013selective,cao2010transfer,zhao2017unified,zhao2013active,zhang2017survey}. Nevertheless, existing MEM approaches usually assume that data distributions/spaces are shared between sources, or sufficient annotations of entity correspondence between sources can be acquired~\cite{chen2020multi,lu2013selective,cao2010transfer,zhao2017unified,zhao2013active}, which may not hold in real-applications. Consider an example of two search systems of two countries, respectively, where query words, users, items can be regarded as three types of entities. However, 
users and items may not share between countries and may have different feature spaces. And the query words and item descriptions may be in different languages. 

\begin{figure}[h]
  \centering
  \includegraphics[width=\linewidth,height=2.2cm]{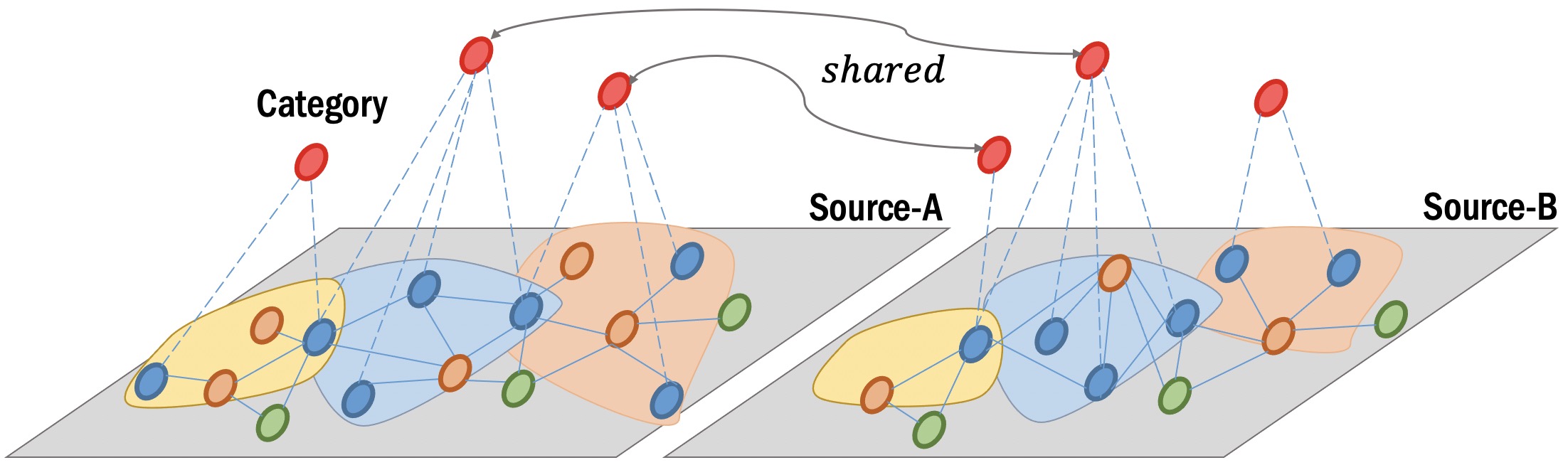}
  \caption{An example of our proposed~\problemname~problem. Circles with different colors denote different types of entities. Lines that connect entities denote annotated relationships. Top circles represent categories. The categories connected by curves are shared between sources. Best view in color.}
  \label{fig:problem}
\end{figure}

Therefore, 
this paper relaxes the above assumptions and proposes a MEM setting in that data distributions/spaces are not shared between sources, and cross-source entity-correspondence is not provided, while the matching tasks of different sources are similar.
In addition, we also assume that the shared categories of entities between sources are provided. As in the above example, two items from different countries may share the same item category (e.g., both items are books). In real applications, such coarse correspondences are usually easy to obtain and possible to be exploited to generate high values.
To sum up, this problem will be referred to as a Relaxed Multi-source Large-scale Entity-matching (\problemname) problem, which is illustrated in Fig.~\ref{fig:problem}.

In this paper, our approach to~\problemname~is via Graph Neural Network (GNN)~\cite{wu2020comprehensive} 
(with entities being represented as nodes and matching relationships represented as weighted edges), for its effectiveness in solving the data-sparsity problem~\cite{chen2020multi,zhang2019oag,wang2020m2grl,zhang2018link,ying2018graph,kipf2016semi,grover2016node2vec,zang2020moflow}, because indirect connections can complete sparse direct connections.
Specifically, each source is modeled as a graph. Here we state two major challenges for GNN in handling the \problemname~problem.


The first challenge is \emph{effective and efficient information sharing} between cross-source large graphs.
First, besides sharing models, node alignment between graphs is necessary because of differences in data distribution or feature spaces. Since no fine-grained correspondence is provided in each category, unsupervised alignment is required, which is well formulated and handled by graph matching approaches~\cite{yan2016short,xu2019gromov,chen2020graph}. These approaches can align nodes based on consistency between node representations and node-node matching relationships. For efficient aligning large-scale entities, one can adopt sliced graph matching approaches~\cite{rabin2011wasserstein,titouan2019sliced}, which have a time complexity of $O(nlogn)$. However, another challenge,  \emph{negative transfer}~\cite{pan2009survey}, still stands in the way, which is referred to the accuracy being degraded due to involving auxiliary data sources for training. Recent Pareto Multi-Task Learning (PMTL) approaches~\cite{lin2019pareto,mahapatra2020multi,shah2016pareto,navon2020learning,lin2020controllable,ruchte2021efficient,deist2021multi} are promising to constrain negative transfer. However, for deep neural networks, constraining negative transfer for one source may still result in poorly-learned node representations for other sources, which leads to the \emph{entanglement of both challenges}: poorly-learned node representations could decrease the accuracy of unsupervised node alignment, and incorrect node alignment may generate worse node representations for other sources during constraining negative transfer for one source.  


To tackle the entanglement of both challenges, we propose a 
an incentive compatible~\cite{roughgarden2010algorithmic} mechanism to render every target source able to optimize its model based on its true preference without worrying about deteriorating representations of other sources. This mechanism first optimizes the Pareto front by optimizing information sharing, then fix information sharing and mitigate negative transfer for target sources. This mechanism is supported by our conclusion that the entanglement of both challenges results from a deteriorating Pareto front, which is generated by our discussion about the relationship between negative transfer, Pareto front, and cross-source alignment.
In our discussion, to explicitly control the negative transfer in the training process, we define a Training Negative Transfer (TNT). We show that the Pareto front depicts the lower bound of TNT for each target source conditioning on certain improvements of other sources. And Pareto fronts with large Hypervolumes Under the Front (HUF) or concave shapes may have large lower bounds of negative transfer. We point out that information sharing plays a critical role to form the Pareto front. 
Because the Pareto front results from conflicts between the objectives, and incorrect sharing may increase conflicts. 




\begin{figure*}[t!]
  \centering
  \includegraphics[width=0.7\linewidth]{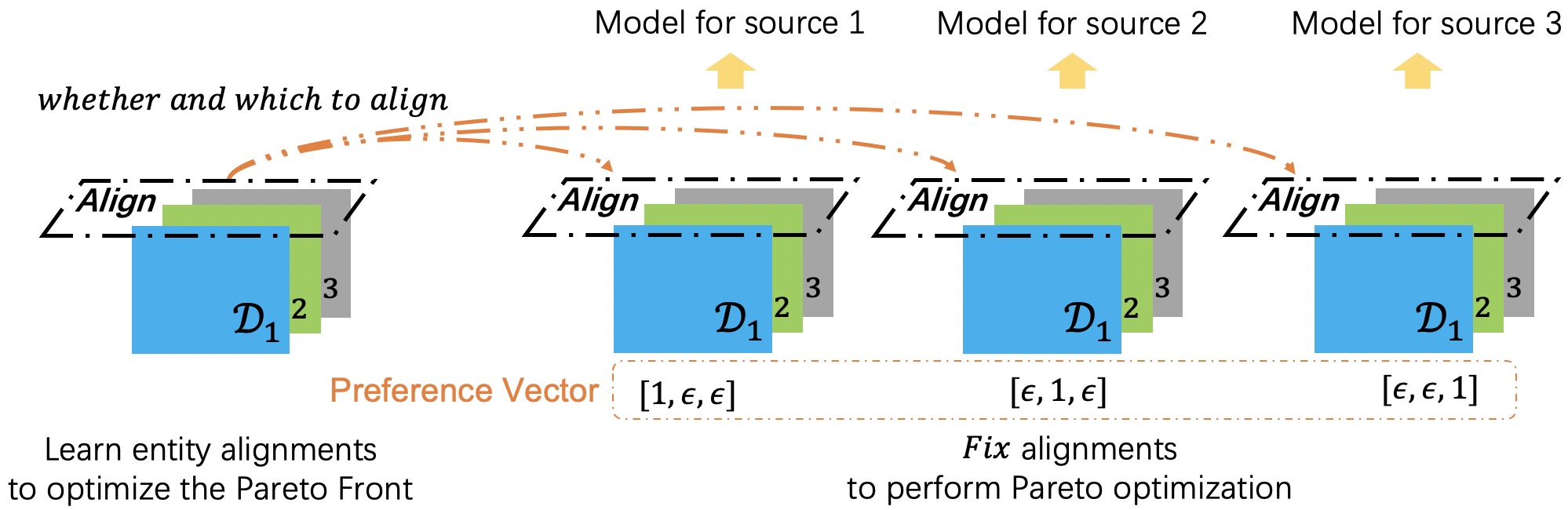}
  \caption{The architecture of our proposed framework. $\epsilon\geq0$ is a small scalar close to zero. Best view in color.
  }
  \label{fig:framwork}
\end{figure*}

According to the proposed mechanism, we propose an Incentive Compatible Pareto Alignment (\methodname) method for the \problemname~problem. The architecture of our proposed framework is presented in Figure~\ref{fig:framwork}. For learnable information sharing, we establish a node alignment model to learn whether and which to align based on sliced graph matching in each category. For front optimization, to minimize the HUF and encourage a convex front, we optimize by the convex combination of objectives. After the alignments are optimized, constrained by the fixed alignments, we perform Pareto optimization (with a nearly one-hot preference vector) to mitigate negative transfer for each source.

We release an International Entity Graph (IEG) dataset, which is collected from traffic logs
of our online search system, and contains data from six countries.
Besides the IEG dataset, we evaluate our \methodname~method on three real-world large-scale benchmark datasets.
We also conduct online A/B test experiments at a search advertising platform. Comprehensive empirical evaluation results demonstrate the effectiveness and superiority of our proposed method. 
 Our code is available online at {\url{https://github.com/anonMLresearcher/ICPA}}.

 


\section{Related Works}\label{sec:related_work}

Multi-Objective Optimization (MOO) refers to the paradigm of learning multiple related objectives together~\cite{boyd2004convex,miettinen2012nonlinear,caruana1997multitask,zhang2017survey}. 
Recently, there are four thrusts of MOO methods that consider the negative transfer problem. The first family resorts to exploiting task relatedness~\cite{wang2019characterizing,zhao2019multiple,feng2020kd3a,mao2020adaptive,ma2020adaptive,wang2019learning}. However, they end up highlighting the data sources with more consensus knowledge, which still cannot rigorously define and constrain the accuracy drop for each source in their learning objectives. The second family learns separate models for shared information and specific information, respectively~\cite{ma2018modeling,park2019soft,peng2020improving,chen2020boosting,tang2020progressive}. However, the negative transfer may still happen in the models which learn shared information, and is not well defined or formulated in these works. The third family is based on meta-learning~\cite{finn2017model,li2018learning} which conducts a ``learning source A to learn source B'' paradigm to constrain the negative transfer between two sources. However, this paradigm only points out the learning destination (e.g., source B), but still does not constrain accuracy sacrifice. 
While the fourth family, PMTL~\cite{lin2019pareto,mahapatra2020multi,shah2016pareto,navon2020learning,lin2020controllable,ruchte2021efficient,deist2021multi}, rigorously defines that a Pareto improvement improves the objective on source A but does not sacrifice the objective on source B, 
and then is promising to constrain negative transfer. Nevertheless, none of these methods considers optimizing entity alignment. In this paper, we discuss mitigating negative transfer while performing cross-source alignment, and handle the entanglement between both challenges.

Graph matching~\cite{yan2016short} learns an optimal correspondence between the nodes of multiple graphs in an unsupervised manner, which is based on an optimal alignment of information from nodes, edges, or higher-order topological structures. Although it is an unsupervised method, it has achieved successes in multi-modal learning~\cite{chen2020graph}, neural language processing~\cite{xu2019cross}, and 3D shape correspondence~\cite{halimi2019unsupervised,maron2018probably}. 
Among graph matching approaches, it is common to perform alignment by Wasserstein distance (WD)~\cite{peyre2019computational}, Gromov-Wasserstein distance~\cite{peyre2016gromov}, and them both~\cite{xu2019gromov,chen2020graph}. 
Recent graph-matching studies combine WD and GWD, and learn the shared correspondence between WD and GWD, for improving effectiveness~\cite{xu2019gromov,chen2020graph}. However, these methods have a relatively high computational complexity of $O(n^3)$, and they are not scalable for large graphs.
On the other hand, for scalability, Xu~\emph{et al.}~\cite{xu2019scalable} developed a graph-partition based method, Rabin~\emph{et al.}~\cite{rabin2011wasserstein} proposed a sliced WD (SWD) method and Titouan~\emph{et al.}~\cite{titouan2019sliced} proposed a sliced GWD (SGWD) method. These methods can reduce the complexity to $O(nlogn)$. However, these methods do not consider 
mitigating negative transfer resulted from incorrect alignment, nor whether a node or edge is allowed to align.





\section{Methodology}\label{sec:method}

In this section, we present the detailed methodology of our method. First, we define the notations and problem settings of our study. In this paper, we denote $[k']$ as the index set $\{1,2,\ldots,k'\}$ .

 Consider a dataset $\mathcal{G}=\{\mathcal{D}_1,\ldots,\mathcal{D}_m\}$ consisting of $m$ data sources. For the $j$th data source,
$\mathcal{D}_j=\{\mathcal{V}_j,\mathcal{E}_j,\mathcal{C}_j\}$ is a graph consisting of its nodes $\mathcal{V}_j$ ,edges $\mathcal{E}_j$, and categories $\mathcal{C}_j$. For each $j\in[m]$, the nodes $\mathcal{V}_j=\{\v_j^i\}$ are allowed to be from different types,
where $\v_j^i$ represents the learned representation of the $i$th node. Each node may have its own features, especially its unique features from its identity. $\mathcal{E}_j=\{(i,l)\}$ collects undirected edges between nodes.
The categories $\mathcal{C}_j=\{c_j^i\}$ include the category of each node, where $c_j^i\in[K]$.
Considering the~\problemname~problem, feature spaces and feature distributions of nodes are not shared between different sources. Identity correspondences of nodes are not given either. Whereas the categories are shared between sources.

The task of the MEM in this paper is defined as: for $j\in[m]$, when considering the $j$th data source as the target source, train a shared GNN on the entire training dataset $\mathcal{G}$, predict whether two nodes are connected with an edge on a testing dataset $\calD'_j$ which is i.i.d. with $\calD_j$, and maximize the edge prediction accuracy on $\calD'_j$. The prediction accuracy on the testing set will be referred to as the generalization performance.


\subsection{Discussions of Training Negative Transfer, Pareto Front, and Information Sharing}\label{subsec:discuss}

In this section, we discuss the following questions: 1) What is the relationship between training negative transfer and the Pareto front of multi-source objectives? 
2) How to optimize the Pareto front to mitigate training negative transfer? 3) For a fixed Pareto front, how to improve generalization performance by cooperating with other objectives while guaranteeing no training negative transfer? 4) What is the relationship between the Pareto front and information sharing? The proofs of theoretical results are deferred to the supplementary material.

\textbf{Q1: What is the relationship between training negative transfer and the Pareto front of multi-source objectives?}
 
First, for multi-source objectives, we denote by $\calL_1,\ldots,\calL_m$ the edge prediction objective functions for all the sources and evaluated on $\calD_1,\ldots,\calD_m$, respectively. The objectives can take arbitrary forms, the smaller the better. 
Let $\nnu^0=[\nu_1^0,\ldots,\nu_m^0]$ such that $\nu_j=\min_{f\in\mathcal{F}|\mathcal{A}}\calL_j(f)$ for each $j\in[m]$, where $\mathcal{F}|\mathcal{A}$ denotes the model space $\mathcal{F}$ constrained by a specific algorithm $\mathcal{A}$.

\begin{figure}[t]
\centering
\subfigure[]{\includegraphics[width=0.4\linewidth]{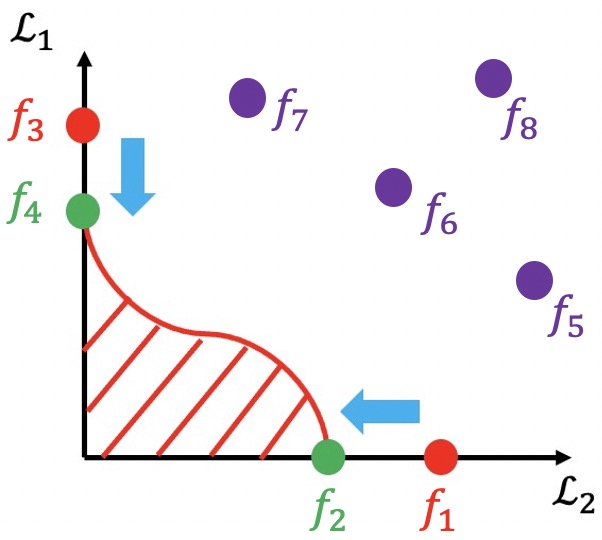}}
\ \ \ \ \ \ \ \ \ \ 
\subfigure[]{\includegraphics[width=0.5\linewidth]{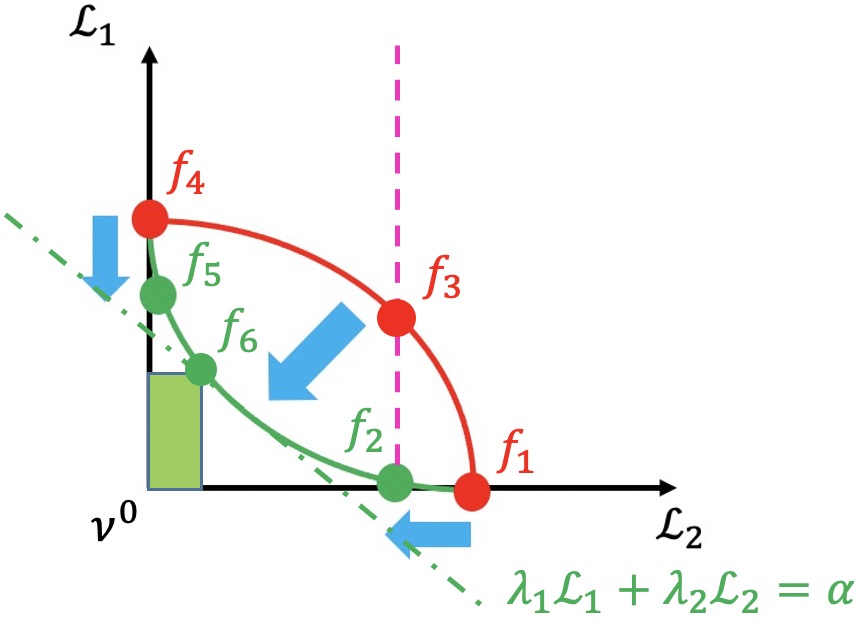}}
\caption{Examples of negative transfer and Pareto fronts. Each point represents a trained model. Curves are the Pareto fronts. Best view in color.}
\label{fig:front}
\end{figure}

Then, the training negative transfer for~\problemname~is defined below, following Wang~\emph{et al.}~\cite{wang2019characterizing}.
We provide examples of TNT in Fig.~\ref{fig:front} (a), where $f_1$ and $f_2$ have no TNT for $\calL_1$ because they achieve the global minimum of $\calL_1$. Whereas  $f_3,f_4,\ldots,f_8$ have TNT for $\calL_1$.
\begin{definition}[Training Negative Transfer (TNT)]\label{df:negative_transfer_train}
Given an algorithm $\calA$, for source $j\in[m]$ and a learned model $f=\calA(\calG)$, 
the degree of training negative transfer is defined by
\begin{equation}\label{eq:criterion_train_amount}
\varepsilon_j(f)=\calL_j(\calA(\calG))-\calL_j(\calA(\calD_j))=\calL_j(f)-\nu_j^0(f).
\end{equation} 
 
\end{definition}



The definition of Pareto front~\cite{zitzler1999multiobjective} is involved in the following.
An example of the Pareto front is provided in Fig.~\ref{fig:front} (a), where the red curve is the Pareto front. No model exists under the front because of the conflicts between the objectives.
\begin{definition}[Pareto Front~\cite{zitzler1999multiobjective}]\label{df:pareto_optimal}
For two models $f_1$ and $f_2$, we write $f_1\prec f_2$ 
if and only if there exists some $j\in[m]$ such that $\calL_j(f_1)<\calL_j(f_2)$, and for all other $j'\in[m],j'\neq j$, we have $\calL_{j'}(f_1)\leq\calL_{j'}(f_2)$. 
A model $f$ is said to be Pareto optimal if there does not exist a $f'$ such that $f'\prec f$. The set of all Pareto optimal models is named the Pareto front.
\end{definition}

For the relationship between Pareto front and TNT, we prove in Theorem~\ref{th:lowerbound} that a Pareto front depicts lower bounds of TNT for each source, conditioning on certain objective improvements of other sources. For example, in Fig.~\ref{fig:front} (a), consider that the current model is $f_1$, to improve $\calL_2$ of at least $\calL_2(f_1)-\calL_2(f_2)$, the lower bound of TNT for $\calL_1$ is $\calL_1(f_2)-\calL_1(f_1)=0$. Whereas, to improve $\calL_2$ of at least $\calL_2(f_1)-\calL_2(f_4)$, the lower bound of TNT for $\calL_1$ is $\calL_1(f_4)-\calL_1(f_1)>0$. 
\begin{theorem}\label{th:lowerbound}
For each source $j\in[m]$, for current model $f_1$ to get achievable improvement $\Delta \calL_{j'}\geq0$ for every other source $j'\neq j$, denote the lower bound of training negative transfer by
\begin{equation}
   \varepsilon_j^0 = \min_{f\in\{f \in (\mathcal{F}|\mathcal{A}) |\forall j'\neq j,\calL_{j'}(f_1)-\calL_{j'}(f)\geq \Delta \calL_{j'} \}} \varepsilon_j(f),
\end{equation}
there must exist a model $f^*$ on the Pareto front such that $\varepsilon_j(f^*)=\varepsilon_j^0$.
\end{theorem}

\textbf{Q2: How to optimize the Pareto front to mitigate training negative transfer?}


Referring to the common metric for the ROC-curve~\cite{fawcett2006introduction} (on which one may evaluate the false positive rate for certain true positive rate): 1) area under the curve (AUC), and 2) convexity of the curve, we evaluate and optimize a Pareto front w.r.t. TNT by: 1) hypervolume under the front (HUF), and 2) convexity of the front.

Here we define the HUF in the following, for readers to better understand the objective for our optimization. 
\begin{definition}[Hypervolume Under the Pareto Front (HUF)]\label{df:criterion_front}
Let $\nnu=[\nu_1,\ldots,\nu_m]\in \mathbb{R}^m$. We write $\nnu^1\preceq\nnu^2$ if and only if for all $j\in[m]$, $\nu_j^1\leq\nu_j^2$.
Denote the objective value vector by $\calLL=[\calL_1,\ldots,\calL_m]$.
For a Pareto front $\calP$, the hypervolume under the front (HUF) is defined as the integral between $\nnu^0$ and the Pareto front:
\begin{equation}\label{eq:criterion_front}
V(\calP)=\int I(\nnu^0\preceq\nnu)I\biggl(\bigcup_{f\in \calP}(\nnu\preceq \calLL(f))\biggr)  d\nnu,
\end{equation} 
where $I(\cdot)$ is the indicator function.
\end{definition}

We provide an example in Fig.~\ref{fig:front} (b), where considering $f_1$ as the current model, to improve $\calL_2$ of at least $\calL_2(f_1)-\calL_2(f_3)=\calL_2(f_1)-\calL_2(f_2)$,
the lower bound of TNT on source $1$ for the upper front is achieved by $f_3$, whereas for the nether front is achieved by $f_2$. And $f_2$ has significantly smaller TNT than $f_3$ does. As shown in Fig.~\ref{fig:front} (b), a front with a large HUF or concave shape may have large lower bounds of TNT. Also, HUF is a ``global'' property: smaller HUF may result in smaller lower bounds of TNT for most cases; whereas the shape of the front is a ``local'' property: in a local region, a convex shape of the front can lead to smaller lower bounds of TNT. 

\textbf{Q3: For a fixed Pareto front, how to improve generalization performance by cooperating with other objectives while guaranteeing no training negative transfer?}
 
First, for generalization, we define the expected risk of a model $f$ for a target source $j\in[m]$:
\begin{equation}\label{eq:exp_risk}
\calR_j(f)= \mathbb{E}_{\calD_j\sim P(\calD_j)}[\calL_j(f(\calD_j))].
\end{equation} 
Then the generalization performance of $f$ on source $j$ is maximized when $\calR_j(f)$ is minimized.

We propose that for a target source $j\in[m]$, a model $\hat{f}$ on the fixed Pareto front such that $\varepsilon_j(\hat{f})=0$ can improve generalization performance while guaranteeing no TNT. We provide an example in Fig.~3 (a), where $f_2$ has the same $\calL_1$ value with that of $f_1$, and also has smaller $\calL_2$ value. Then if sources $1$ and $2$ are correlated, $f_2$ may have better generalization performance on source $1$ than $f_1$, i.e., $\calR_1(f_2)<\calR_1(f_1)$. 
Because in this case, part of $\calD_2$ can be regarded as drawn from $P(\calD_1)$. Therefore, optimizing $\calL_2$ can be regarded as partially optimizing $\calR_1$. The existence of such a solution model is guaranteed by the result below.
\begin{theorem}
For each source $j\in[m]$, there must exist a model $f^*$ on the Pareto front such that $\varepsilon_j(f^*)=0$.
\end{theorem}

\textbf{Q4: What is the relationship between the Pareto front and information sharing?}

We point out that information sharing may significantly influence the Pareto front. Because the Pareto front is formed by conflicts between objectives, and incorrect or contradictory information sharing may directly increase conflicts.
For an example of the~\problemname~problem, if we align an entity \texttt{apple} in source $1$ with an entity \texttt{football} in source $2$ and align \texttt{orange} in source $1$ with \texttt{orange} in source $2$, then the objective $\calL_1$ to learn the pair $\langle$\texttt{apple},\texttt{orange}$\rangle$ as \textit{similar} will be contradictory with the objective $\calL_2$ to learn the pair $\langle$\texttt{football},\texttt{orange}$\rangle$ as \textit{dissimilar}. In general, conflicts increase the difficulty to minimize multiple objectives simultaneously, and then worsen the Pareto front (e.g., enlarge the HUF or render it non-convex).

Based on the discussions above, we interpret the entanglement between cross-source alignment and mitigating negative transfer.
At first, one performs unsupervised node alignment based on initialized node representations, which may result in many incorrect alignments. The incorrect alignments may form a non-convex Pareto front with large HUF, which has large lower bounds of TNT. Then, based on the poor front, one mitigates negative transfer for a target source. Due to large lower bounds of TNT, to guarantee the target source having little TNT, other sources may have large TNT, which will result in poor node representations for other sources. Finally, based on poor node representations, the Pareto front in the next round may have larger HUF and tend to be concave. In conclusion, the entanglement results from a deteriorating Pareto front.

\subsection{Learning Whether and Which to Align}

As discussed in Section~\ref{subsec:discuss}, because the aforementioned entanglement of challenges results from a deteriorating Pareto front, we propose to optimize and fix the front. And since the front could be significantly affected by cross-source alignment, we establish a parameterized alignment model to learn how to align. Specifically, we propose the following loss function to learn whether and which to align between every pair of sources $(j,j')$ in each category, where $j,j'\in[m], j\neq j'$. For brevity, we omit the superscripts for denoting sources.
\begin{equation}\label{eq:wd}
   \ell_a(\ppi,\t;\ttheta_n,\ttheta_t) = \sum_{i,l}\pi_{il}t_it_ld(\x_i,\y_l;\ttheta_n) + \ell_g(\t;\ttheta_t),
\end{equation}
where $\ppi\in\Pi(a,b)=\{\ppi\in\mathbb{R}_+^{n_a,n_b}|\sum_l\pi_{i,l}=a_i,\sum_i\pi_{i,l}=b_l\}, \a\in\Sigma_{n_a},\b\in\Sigma_{n_b}$, and for an integer $n$, $\Sigma_n=\{a\in\mathbb{R}_+^n|\sum_i a_i=1\}$. $\x_i$ is the node representation of $i$th node of source $j$, whereas $\y_l$ is the node representation of $l$th node of source $j'$. $\pi_{il}$ denotes \emph{how much probability for $\x_i$ to align $\y_l$}. $d(\cdot,\cdot)$ is a distance or discrepancy function. $t_{\cdot}=G(\v_{\cdot};\ttheta_t)\approx p(\delta=1\mid \v_{\cdot};\ttheta_t)$ denotes the approximate Bernoulli probability out of a gating network $G$ for \emph{whether to align} $ \v_{\cdot}$, where $\v_{\cdot}$ can be $\x_{\cdot}$ or $\y_{\cdot}$. $\ttheta_n$ denotes the parameters to generate node representations, whereas $\ttheta_t$ denotes the parameters of the gating network.

In Eq.~\eqref{eq:wd}, the alignment plan $\ppi$ decides \emph{which nodes in another source to align} for each node. Whereas the gates $t$s learn \emph{whether to align} for each node and can relax the compulsory constraints from $\ppi$ that a node must align some nodes in another source. Therefore, the gates can mitigate negative transfer from incorrect compulsory alignment. The $\ell_g(\t;\ttheta_t)$ denotes the information maximization loss~\cite{MI3} for the gates to prevent the trivial solution of sharing no nodes, and is defined as
\begin{equation}\label{eq:mi}\small
\ell_g(\t;\ttheta_t)=-\sum_it_i\log(t_i)-(1-t_i)\log(1-t_i)+\bar{t}\log(\bar{t})+(1-\bar{t})\log(1-\bar{t}),
\end{equation} 
where $\bar{t}$ is the average of $t$s. This loss encourages individual certainty and global diversity. 

Note that Eq.~\eqref{eq:wd} is a general loss function. In detail, we adopt sliced graph matching techniques~\cite{rabin2011wasserstein,titouan2019sliced} to render the computation of the alignment plan $\ppi$ to have time complexity of $O(nlogn)$, which is deferred to the supplementary material.

\subsection{Optimizing the Pareto Front for Alignment Learning}
 
Based on the discussions in Section~\ref{subsec:discuss}, to optimize the Pareto Front, we should minimize the HUF and encourage the convexity of the front.

For minimizing the HUF, one can adopt recent Pareto front learning methods~\cite{shah2016pareto,navon2020learning,lin2020controllable,deist2021multi}. However, these methods usually require hyper-networks, multiple models, or complicated computations which are relatively not efficient for large-scale entities. Thus, we approximate the HUF minimization by minimizing the hypervolume of the hyperrectangle resulted from each model $f$: $V^{rec}(f)=\prod_j(\calL_j(f)-\nu_j^0)$, which is illustrated by the hyperrectangle in Fig.~\ref{fig:front} (b). And since $\nnu^0$ is usually unknown and minimizing $\prod_j\calL_j(f)$ cannot guarantee minimizing $V^{rec}(f)$. We then minimize the upper bound of $V^{rec}(f)\leq (\frac{1}{m}\sum_j(\calL_j(f)-\nu_j^0))^m$. As $\nnu^0$ is fixed, we minimize $\frac{1}{m}\sum_j\calL_j(f)$---which is the average of objectives. Considering to minimize the HUF of all possible combinations of the objectives, we should minimize the average of each possible combination of the objectives. And further considering all scalings for the objectives, we should minimize each convex combination of the objectives: $\sum_j\lambda_j\calL_j, \forall \llambda \in \{\llambda\in\mathbb{R}_+^m|\sum_j\lambda_j=1\}$.



To encourage the convexity of the front, inspired by the partial converse of the supporting hyperplane theorem~\cite{boyd2004convex}, we propose to encourage every point on the Pareto front to have a supporting hyperplane. We also use the proposed convex combinations of the objectives above to approximate this target, which is illustrated by the dash line in Fig.~\ref{fig:front} (b).


\subsection{Our Framework}\label{subsec:framework}

We summarize our ICPA framework in Algorithm~\ref{alg:ICPL} which is also illustrated in Fig.~\ref{fig:framwork}. We first learn an alignment model by the objective in Eq.~\eqref{eq:opt_front}. The parameters are fixed to learn the final model. Then we perform a Pareto optimization to learn a model for the target source, constrained by the fixed alignments from the optimized alignment model. The choice of the preference vector aims to improve generalization while mitigating TNT. The explanation is deferred to the supplementary material. Since the alignments are optimized and fixed, each target source can optimize its model by PMTL based on its true preference without worrying about deteriorating representations of other sources. Therefore, our method can be regarded as incentive compatible~\cite{roughgarden2010algorithmic} 
for multiple sources.

\begin{algorithm} [H]
\caption{Incentive Compatible Pareto Alignment}
\label{alg:ICPL}{\small
\begin{algorithmic}[1]
\REQUIRE{Data set $\mathcal{G}=\{\mathcal{D}_1,\ldots,\mathcal{D}_m\}$, the target-source index $j^*$. Alignment loss weight $\beta>0$.}
\ENSURE{GNN model parameters: $\hat{\ttheta}$.}
\WHILE{not converge}
\STATE Sample a mini-batch of nodes in the same category, and obtain losses $\calL_1,\ldots,\calL_m$.
\STATE Sample a vector $\llambda\in \{\llambda\in\mathbb{R}_+^m|\sum_j\lambda_j=1\}$.
\STATE Optimize the alignment model:
\begin{equation}\label{eq:opt_front}\small
\begin{split}
     &\ttheta_n^0,\ttheta_t^0,\ppi^0,\t^0 =\\
     &\arg\min_{\ttheta_n,\ttheta_t,\ppi,\t} \ \sum_j\lambda_j\calL_j(\ttheta_n)  + \beta\ell_a(\ppi,\t;\ttheta_n,\ttheta_t),
\end{split}
\end{equation}
where $\ell_a(\ppi,\t;\ttheta_n,\ttheta_t)$ is defined in Eq.~\eqref{eq:wd}.
\ENDWHILE
\STATE Let $\z$ be the preference vector with the $j^*$th element of $1$, and other elements of a small scalar $\epsilon$.
\WHILE{not converge}
\STATE Sample a mini-batch of nodes in the same category, and obtain losses $\calLL=[\calL_1,\ldots,\calL_m]$.
\STATE Get loss weights $\w=\mbox{PMTL}(\calLL,\ttheta_n,\z)$, where PMTL denotes an arbitrary PMTL algorithm.
\STATE Generate note representations $\{\v_i^0\}$ by $\ttheta_n^0$; $\t_i^0=G(\v_i^0;\ttheta_t^0)$ ; $\ppi^0=\min_{\ppi}\ell_a(\ppi,\t^0;\ttheta_n^0,\ttheta_t^0)$.
\STATE Optimize the model for the $j^*$th source:
\begin{equation}\label{eq:opt_single}\small
   \hat{\ttheta}= \arg\min_{\ttheta_n} \ \sum_jw_j\calL_j(\ttheta_n)  + \beta\ell_a(\ppi^0,\t^0;\ttheta_n,\ttheta_t^0).
\end{equation}
\ENDWHILE
\end{algorithmic}}
\end{algorithm}


\noindent\textbf{Time Complexity \;}
Let $B\in\mathbb{Z}_+$ be the batch size, and $D\in\mathbb{Z}_+$ the dimension of model parameters. Since the sliced graph matching methods~\cite{rabin2011wasserstein,titouan2019sliced} have run time $O(BlogBD)$, the state-of-the-art PMTL method~\cite{mahapatra2020multi} has run time $O(m^{2.38}D)$, other processes have run time $O(BD)$, and usually we have $B\gg m$, our method ICPA has total run time of $O(BlogBD+m^{2.38}D+BD)=O(BlogBD)$.

\subsection{Details of the Sliced Graph Matching Technique}

This section introduces the sliced graph matching technique we used for scalable cross-source alignment.
In detail, we use a shared random vector to project all the node embeddings into a shared mono-dimensional space. Then by sorting these 1D values in the shared 1D space. 
 
Denote by $\xxi$ the random vector drawn from a hypersphere such that $\|\xxi\|=1$. The projected 1D nodes are represented as $v_{\cdot}^{\cdot}=\xxi^T{\v}_{\cdot}^{\cdot}$. Then we denote by $v_{\cdot}^{r_i}$ as the $i$th value of $\{v_{\cdot}^{i}\}$ in ascending order. 
Then Eq.~(3) is replaced by 
\begin{equation}\label{eq:align_swd_gated}
\ell_a(\ppi,\t;\ttheta_n,\ttheta_t) =\frac{1}{m(m-1)} \sum_{j\neq j'} \sum_{i} t_{j,j'}^{r_{i}}t_{j',j}^{r_{i}}u(v_j^{r_{i}},v_{j'}^{r_{i}}) + \ell_g(\t;\ttheta_t) ,
\end{equation} 
where $t_{j,j'}^{r_{i}}=p(\delta_{j'}=1\mid {\v}_j^{r_{i}};\ttheta_t)$ denotes the learned Bernoulli probability out of the gating networks for whether to align $ {\v}_j^{r_{i}}$ to source $j'$, whereas $t_{j',j}^{r_{i}}=p(\delta_{j}=1\mid  {\v}_{j'}^{r_{i}};\ttheta_t)$ denotes the learned Bernoulli probability out of the gating networks for whether to align $ {\v}_{j'}^{r_{i}}$ to source $j$.
Intuitively, the sliced technique---projection and sorting---implicitly optimizes the transport plan matrices $\ppi$s to align nodes between sources~\cite{rabin2011wasserstein,titouan2019sliced}. The resulted $\ppi$s can be regarded as binary matrices.
The time complexity is therefore bounded by the run time of the sorting process: $O(nlogn)$.
 
For implementation, we organize the data such that query or item nodes in a mini-batch are basically in the same category, and cover nodes of all the sources. Then all the aforementioned alignments can be performed in a single mini-batch. For each mini-batch, the alignment distances are minimized only when shared categories exist. 
For the case that the numbers of nodes are different between sources, we perform an index interpolation technique. Specifically, assuming the numbers of nodes are required to be $1024$, we first generate an integer indices list from $1$ to $1024$, then we divide each index by $1024$, multiply the actual number of nodes for each data source, and finally round the values as new indices. Then the generated indices have the length of $1024$ and can be used to select nodes in each data source.

		
		
		
		

\section{Experiments}\label{sec:exp}


We evaluate our method on large-scale recommendation and searching datasets. For the former, we conduct experiments on the Amazon-UserBehavior datasets; whereas, for the latter, we conduct experiments on the AliExpress dataset and our released IEG dataset. The detailed re-organization process for each data will be introduced in the following sections. For each method on each dataset, we repeat it $5$ times and report the averaged results. Due to limited space, the detailed results with error bars are deferred to the supplementary material.

\noindent\textbf{Methods for Comparison \;}
We compare our method (Ours) with the state-of-the-art approaches for multi-source entity-matching: MMoE~\cite{ma2018modeling} and M2GRL~\cite{wang2020m2grl}, and GNN baselines: DGI~\cite{velickovic2019deep} and GraphSAGE (GSAGE)~\cite{hamilton2017inductive}.
Specifically, we establish our backbone model (Base) based on GSAGE, where we sample positive node-pairs via the Node2vec~\cite{grover2016node2vec} technique. For MMoE and M2GRL, Base also serves as the backbone models. We also compare the single objective optimization (SOO) which learns on each single data source only, using the Base method.
  

\noindent\textbf{Evaluation Metrics \;}
We evaluate by entity matching performance. 
For the AE dataset, we follow Peng~\emph{et al.}~\cite{peng2020improving} to use Normalized
Discounted Cumulative Gain (NDCG) @k for evaluation with $k=17$, which is defined as the NDCG measured on the items with top $k$ matching scores for a user and is averaged over all the users.
For the AU dataset, we follow Zhu~\emph{et al.}~\cite{zhu2019joint} to use the F-Measure@k for evaluation with $k=200$, which is averaged over all the users.  Finally, for the IEG dataset, we evaluate by F-Measure@k with $k=50$, weighted averaging over all the queries.
 


\noindent\textbf{Data Preprocessing \;}
Following the practise of M2GRL~\cite{wang2020m2grl}, for constructing item-item edges, we first order items with user behavior (e.g., clicks) for each user/query by the timestamps. Then we add an edge between two items if the number of items between them is less than $9$. Finally, we set the weight of each edge by the occurrence frequency of the edge.

\noindent\textbf{Implementation Details \;}
We learn node feature by weighted binary classification, where for each node, we sample $1$ neighbor node via a random walk as the positive node and sample $6$ nodes in the same category with the positive node with sampling probabilities proportional to their degrees. The weight for the positive nodes is $2$. For the random walk, the return parameter $p$ and the in-out parameter $q$ are both $1$. For graph convolution, for each node, we sample $5$ neighbor nodes for aggregate.
We adopt average pooling for aggregating the same type of nodes and concatenation for combine different types of nodes. 
We sample $\llambda$ from the Dirichlet distribution.
For Pareto optimization, we adopt ParetoMTL~\cite{lin2019pareto}. For inference, we adopt sigmoid after the node-representation inner-product as the matching probability. The batch size is $1024$. 
We equal the model capacity of each method for fair comparisons.  Our implementation uses Tensorflow~\cite{abadi2016tensorflow}. We run each method on our cluster with $10$ instances (each has $24$ CPU cores). Other details are in the supplementary material.

\subsection{AliExpress (AE)}\label{subsec:ae}

The AE dataset~\cite{peng2020improving} is collected and sampled from traffic logs of the AliExpress search system and consists of data from 5 countries: Spain (ES), French (FR), Netherlands (NL), Russia (RU), and America (US). The datasets are organized in a user-item interaction form, and contain specific features for users and items, respectively. We contact the authors of AE and acquire the shared 252 categories between countries. 
We treat each user-item click behavior as an edge with a weight of $1$. Each country is treated as a source. There is no entity-correspondence between sources. The statistics of AE are listed in Table~\ref{tab:stat_ae}. Since AE is for the click-through rate prediction, we directly use the binary click labels for edge prediction.
The training and testing sets are separated according to~\cite{peng2020improving}.



\setlength{\tabcolsep}{2pt}
\begin{table}

  \caption{Statistics of the AE dataset.}
  \centering
  \label{tab:stat_ae}
  \begin{tabular}{lcccc}
    \toprule
    Source & \#edges & \#users & \#items \\
    \midrule
    ES &	0.8M &	1.7M &	31M\\
    FR &	0.5M &	1.4M &	26M\\
    NL &	0.3M &	1M &	17M\\
    RU &	3.6M &	7.4M &	129M\\
    US &	0.4M &	1.5M &	27M\\
  \bottomrule
\end{tabular}
\end{table}

\setlength{\tabcolsep}{2pt}
\begin{table}
  \caption{Results of NDCG scores on AE.}
   \centering
  \label{tab:results_ae}
  \begin{tabular}{lccccc}
    \toprule 
    Method  & ES & FR & NL & RU & US \\
    \midrule
    SOO   &  43.34\%  & 40.40\% & 39.88\% & 41.65\% & 44.07\%  \\
    Base   &  44.49\%&42.92\%&44.56\%&45.19\%&43.62\% \\
    GSAGE &  42.53\% & 41.26\% & 37.65\%  & 42.88\% & 41.67\%     \\
    DGI   & 42.87\%  & 43.19\%  & 41.76\% &  46.33\%  & 42.98\%  \\
    MMoE   &  45.24\%&43.13\%&45.25\%&46.02\%&43.31\% \\
    M2GRL  &  44.10\%&42.56\%&44.38\%&44.78\%&43.23\% \\
    Ours    &  \textbf{46.95\%}&\textbf{45.47\%}&\textbf{46.50\%}& \textbf{48.03\%}&\textbf{47.27\%} \\
  \bottomrule
\end{tabular}
\end{table}

As shown in Table~\ref{tab:results_ae}, our method significantly outperforms the baseline methods, which demonstrates that the effectiveness and superiority of our method are significant.  Moreover, our method does not show negative transfer compared to the SOO method. Whereas the baselines admit negative transfer in some countries. For example, compared with SOO, Base underperforms on US; GSAGE underperforms on ES, NL, and US; DGI underperforms on ES and US; both MMoE and M2GRL underperform on US. All these baselines achieve the relatively highest scores on RU  whose data size is the largest. These phenomena suggest that the baselines will favor the data sources with relatively sufficient edges and then cause negative transfer on those with sparse data. Specifically, because MMoE and M2GRL own a specific model for each source, i.e., specific gates of MMoE and source-wise uncertainty learning of M2GRL, both methods achieve better performances and less negative transfer. Nonetheless, enjoying optimized alignment and negative-transfer-minimized weighting, our method has the best performances and no negative transfer.


\noindent\textbf{Ablation Study \;}
We evaluate three variants of our method by ablating our three components, \emph{i.e.,}
cross-source alignment (Align), alignment learning by front optimization (Front), and Pareto learning based on optimized alignments (Pareto). 
 In Table~\ref{tab:results_ab_ae}, we show the effectiveness of each component. Note that in the scenario without Pareto learning (using fixed linear weighting), the results are close to the best ones, suggesting that our method can improve the front such that linearly weighted objectives can also achieve good performance. In contrast, the variant without front optimization (directly performs alignment and PMTL) suffers severe negative transfer due to the deteriorating front.
 
 
 
\begin{figure}
\centering
 \subfigure{\includegraphics[width=0.8\linewidth]{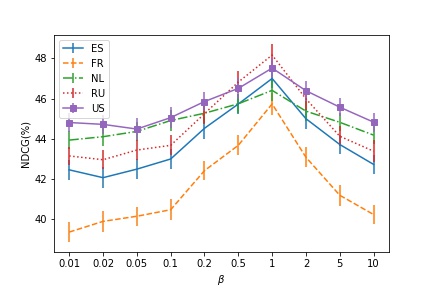}}
\caption{Results of different alignment loss weights on the AE dataset. Best view in color.}
\label{fig:param}
\end{figure}

 
\noindent\textbf{Parameter Sensitivity \;}
 We analyze the parameter sensitivity of the alignment loss weight $\beta$ on the AE dataset. As shown in Fig.~\ref{fig:param}, $\beta$ achieves the optimum around $1$. When $\beta$ is too small, the alignment is weak and limits information sharing. While when $\beta$ is too large, the strong alignment constraints may cause negative transfer.


\setlength{\tabcolsep}{2pt}
\begin{table}

 \caption{Results of the ablation study on AE.}
   \centering
  \label{tab:results_ab_ae}
  \begin{tabular}{lccccc}
    \toprule
    Method  & ES & FR & NL & RU & US  \\
    \midrule
    Ours &  \textbf{46.95\%}&\textbf{45.47\%}&\textbf{46.50\%}& \textbf{48.03\%}&\textbf{47.27\%} \\
     w/o Align &  42.11\%&39.45\%&44.15\%&42.80\%&44.46\%  \\
    w/o Pareto    &  45.74\%&44.13\%&45.65\%&45.92\%&46.18\%  \\
    w/o Front  & 39.89\%&39.42\%&44.41\%&43.25\%&41.28\%   \\
  \bottomrule
\end{tabular}
\end{table}

\subsection{Amazon-UserBehavior (AU)}\label{subsec:au}
 
The Amazon dataset~\cite{he2016ups} consists of product reviews from Amazon users, while the UserBehavior dataset~\cite{zhu2019joint} is a subset of Taobao user behavior data. Both datasets are organized in a user-item interaction form and have category annotations for each item. 
We treat each user-item (u-i) interaction as an edge with a weight of $1$. We also refer to~\cite{wang2020m2grl} to construct item-item (i-i) edges when neighboring in time with a time window of $9$.  Each dataset is treated as a source. We contacted the authors of~\cite{zhu2019joint} and acquired the root category for each item, and established $20$ shared categories between sources. There are no entity-correspondence, nor other shared features between sources. The statistics of AU are listed in Table~\ref{tab:stat_au}. The training, validation, and testing sets are separated according to~\cite{zhu2019joint}. 



\setlength{\tabcolsep}{2pt}
\begin{table}
  \caption{Statistics of the AU dataset.}
  \label{tab:stat_au}
   \centering
  \begin{tabular}{lcccc}
    \toprule
    Source & \#u-i edges & \#i-i edges & \#users & \#items \\
    \midrule
    Amazon   & 80M &   285M  & 20M & 8M    \\
    UserBehavior   & 100M & 533M & 1M & 4M  \\
  \bottomrule
\end{tabular}
\end{table}
\setlength{\tabcolsep}{2pt}
\begin{table}
  \caption{Results of F-measure scores on AU.}
  \label{tab:results_au}
   \centering
  \begin{tabular}{lcc}
    \toprule
    Method  & Amazon & UserBehavior \\
    \midrule
    SOO   & 1.80\%  &  5.17\%  \\
    Base   & 1.47\%  & 9.37\%   \\
    GSAGE & 1.16\%     & 6.54\%   \\
    DGI   &  1.57\%  & 8.63\%   \\
    MMoE   & 1.40\%   &   9.72\%  \\
    M2GRL  & 1.30\%   & 7.35\% \\
    Ours    & \textbf{2.43\%}&\textbf{10.21\%}  \\
  \bottomrule
\end{tabular}
\end{table}

For evaluation on the AU dataset, the user-behavior sequence data used by Zhu~\emph{et al.}~\cite{zhu2019joint} are not adopted for brevity. Thus, for each user, the first half items along the timeline used as a known user-behavior sequence by Zhu~\emph{et al.} is exploited for matching the second-half items using the following equation:
\begin{equation}\label{eq:au_eval}
\begin{split}
    p(\y\mid\x)&=\sum_{\y'}p(\y,\y'\mid\x)=\sum_{\y'}p(\y'\mid\x)p(\y\mid\x,\y')\\
    &\approx\sum_{\y'}p(\y'\mid\x)p(\y\mid \y')=\sum_{\y'}\frac{p(\y',x)}{\sum_{\y''}p(\y'',x)}p(\y\mid \y'),
\end{split}
\end{equation} 
where $\x$ denotes an arbitrary user, $\y',\y''$ denote items in the first half items along the timeline, and $\y$ denotes an item in the second half items along the timeline. $p(\y',x)$ and $p(\y'',x)$ denote the edge weights between the enclosed nodes, respectively. $p(\y\mid \y')$ denotes the matching probability out of node embedding between $\y$ and $\y'$.
 
As shown in Table~\ref{tab:results_au}, our method outperforms the baseline methods, especially on the Amazon dataset whose sparseness is more significant, which demonstrates that the effectiveness of our method to handle data sparsity. Note that AU is difficult because nothing is shared between sources except a few coarse categories, plus feature information is very limited. Therefore, MMoE and M2GRL cannot unleash their strength to tackle negative transfer and show relatively low performances on Amazon.
DGI performs better on Amazon because its learned structure information can mitigate the information deficiency. Whereas our method performs better, suggesting the optimized alignment is more important for information complement.

\subsection{International Entity Graphs (IEG)}\label{subsec:ieg}

Our released IEG dataset is collected and sampled from traffic logs of our online search system and consists of data from 6 countries: Indonesia (ID), Malaysia (MY), Philippines (PH), Singapore (SG), and Thailand (TH), and Vietnam (VN). 
The datasets are organized in a query-item interaction form. Each interaction summarizes the number of clicks. There are 4273 shared categories between countries. IEG contains specific features for queries and items respectively. These features are all discrete and include many identity features, therefore, are significantly different between countries. We treat each query-item (q-i) pair as an edge with the weight of the number of clicks. We also use the same procedure as in Section~\ref{subsec:au} to construct item-item edges. Each country is treated as a data source. There are 16 million item-correspondence between TH and PH. The statistics of IEG are listed in Table~\ref{tab:stat_ieg}. The first 31 days and the last 7 days of the data are treated as training and testing sets, respectively. The frequency of each testing query is provided for weighting.  



\begin{table}
 \caption{Statistics of the IEG dataset.}
  \label{tab:stat_ieg}
   \centering
  \begin{tabular}{lcccc}
    \toprule
    Source & \#q-i edges & \#i-i edges & \#queries & \#items \\
    \midrule
    ID &	19M & 132M &	1.9M &	1.7M\\
    MY &	0.2M & 13M &	0.3M &	0.7M\\
    PH &	6M & 139M &	0.7M &	3.2M\\
    SG &	0.5M & 3M &	0.1M &	0.3M\\
    TH &	5M & 111M &	0.5M &	2.9M\\
    VN &	2.3M & 15M &	0.3M &	0.6M\\
  \bottomrule
\end{tabular}
\end{table}

\setlength{\tabcolsep}{2pt}
\begin{table}
 \caption{Results of F-measure scores on IEG.}
  \label{tab:results_ieg}
   \centering
  \begin{tabular}{lcccccc}
    \toprule 
    Method  & ID & MY & PH & SG & TH & VN \\
    \midrule
    SOO   &  22.46\%  &15.93\% &18.48\% &9.67\% &20.17\% &16.53\%  \\
    Base   & 22.87\% &16.12\% &19.39\% &9.64\% &21.30\% &17.64\% \\
    GSAGE   &22.79\% &15.75\% &18.95\% &9.26\% &20.77\% &17.21\% \\
    DGI   &  23.82\% &16.19\% &18.89\% &9.46\% &21.06\% &20.37\%  \\
     MMoE   &  23.12\% &16.71\% &19.37\% &9.60\% &21.33\% &20.65\%   \\
    M2GRL  &  22.98\% &16.13\% &18.75\% &9.69\% &20.72\% &17.12\%   \\
      Ours    & \textbf{24.17\%}&\textbf{18.34\%}&\textbf{20.10\%}&\textbf{12.60\%}&\textbf{22.29\%}&\textbf{21.67\%}   \\
  \bottomrule
\end{tabular}
\end{table}
\setlength{\tabcolsep}{2pt}
\begin{table}
  \caption{Cases of irrelevant query-item pairs.}
  \label{tab:case}
  \centering
  \begin{tabular}{ll}
    \toprule
    query & item \\
    \midrule
   Inflatable water bath &	3-meter 3-story pool  \\
   Shedding spray &	Deodorant Body Spray  \\
   Learning equipment &	Laminating photo card \\
  \bottomrule
\end{tabular}
\end{table}


\setlength{\tabcolsep}{2pt}
\begin{table}[h]\small
  \caption{Online A/B test results in our system.}
  \centering
  \label{tab:abtest}
  \begin{tabular}{lcccc}
    \toprule 
    Country  & Revenue & \#Clicks & RPM & CTR \\
    \midrule
    ID&	1.16\%&	0.99\%&	0.78\%&	0.61\% \\
MY&	3.16\%&	1.23\%&	2.19\%&	0.27\% \\
PH&	2.52\%&	1.12\%&	1.27\%&	0.22\% \\
SG&	5.65\%&	4.87\%&	4.61\%&	3.84\% \\
TH&	1.69\%&	0.94\%&	0.93\%&	0.18\% \\
VN&	1.38\%&	0.99\%&	0.73\%&	0.34\% \\
ALL&	2.05\%&	1.06\%&	1.50\%&	0.50\% \\
  \bottomrule
\end{tabular}
\end{table}


As shown in Table~\ref{tab:results_ieg}, our method still outperforms the baseline methods, especially on MY and SG whose data sizes are relatively small, which demonstrates the effectiveness of our Pareto learning module and the superiority of our optimized alignment module. For details, no negative transfer is witnessed on PH and TH because of added links. Nonetheless,
there are still some baselines that show negative transfer on SG whose data is sparse. 

\noindent\textbf{Case Study \;}
We sampled specific examples to see how our method works. Table~\ref{tab:case} shows the query-item pairs on TH that are correctly recognized by our method as irrelevant and incorrectly recognized by the direct method to perform alignment and PMTL simultaneously, which suggests that our method can handle the entangled challenges and make a better prediction.

\noindent\textbf{Online A/B Test\;}
We deploy our method for query-item retrieval in our online advertising system and conduct strict online A/B testing experiments. Our method achieves 2.05\% growth on revenue and	 1.06\% growth on clicks. Other detailed results are shown in Table~\ref{tab:abtest}, where
\begin{equation}
     \mbox{RPM}=\frac{\mbox{\# Revenue }}{\mbox{\# Impressions}}, \ 
     \mbox{CTR}=\frac{\mbox{\# Clicks }}{\mbox{\# Impressions}}.
\end{equation}
The click number score evaluates the ability to retrieve items that users are interested to click;
the CTR score evaluates the accuracy of a search system on predicting whether a user will click an impressed item based on the query.
Therefore, the results in Table~\ref{tab:abtest} significantly demonstrate the effectiveness of our method to retrieve relevant items for each query, since 0.1\% improvements in industrial applications are considerable.


\section{Conclusion}\label{sec:conclusion}

In this paper, we investigate the proposed Relaxed Multi-source Large-scale Entity-matching (\problemname) problem in the favored direction of graph-based learning and handle the entanglement between cross-source alignment and mitigating negative transfer. 
We propose an Incentive Compatible Pareto Alignment (\methodname) framework for~\problemname, which renders each source can learn based on its true preference without worrying about deteriorating representations of other sources.
Our experimental results reveal that \methodname~can effectively handle the entangled challenges and demonstrate the superiority of our method. Extensions of our work without coarse correspondence will be explored.




\bibliographystyle{ACM-Reference-Format}
\bibliography{sample-base}

\appendix
\section*{Supplementary Material}

\section{Proofs}
\subsection{Proof of Theorem 1}

\begin{proof}
Without loss of generality, we consider source $1$ as the target source.
Constraining on the conditions: $\forall j'\neq 1,\calL_{j'}(f_1)-\calL_{j'}(f)\geq \Delta \calL_{j'}$, we find the conditional optimum of $\calL_1$:
\begin{equation}
    \nu_1=\min_{f\in\{f\in(\mathcal{F}|\mathcal{A})\mid \forall j'\neq 1,\calL_{j'}(f_1)-\calL_{j'}(f)\geq \Delta \calL_{j'}\}}\calL_1(f).
\end{equation}
Then constraining on $\calL_1(f)=\nu_1$ and the conditions that $\forall j'\neq 1,\calL_{j'}(f_1)-\calL_{j'}(f)\geq \Delta \calL_{j'}$, we find the conditional optimum of $\calL_2$:
\begin{equation}
    \nu_2=\min_{f\in\{f\in(\mathcal{F}|\mathcal{A})\mid \calL_1(f)=\nu_1, \forall j'\neq 1,\calL_{j'}(f_1)-\calL_{j'}(f)\geq \Delta \calL_{j'}\}}\calL_2(f).
\end{equation}
Note that, based on the constraints, we have $\calL_{2}(f_1)-\nu_2\geq \Delta \calL_{2}$.
Then, constraining on $\calL_1(f)=\nu_1,\calL_2(f)=\nu_2$ and the conditions: $\forall j'\neq 1,\calL_{j'}(f_1)-\calL_{j'}(f)\geq \Delta \calL_{j'}$, we find the conditional optimum of $\calL_3$:
\begin{equation}
    \nu_3=\min_{f\in\{f\in(\mathcal{F}|\mathcal{A})\mid \mathcal{T}_{2}, \forall j'\neq 1,\calL_{j'}(f_1)-\calL_{j'}(f)\geq \Delta \calL_{j'}\}}\calL_3(f),
\end{equation}
where event $\mathcal{T}_{2}=\{\forall k\in[2],\calL_k(f)=\nu_k\}$.

Continuing this process until source $m$, we get the final model:
\begin{equation}
    f^*=\arg\min_{f\in\{f\in(\mathcal{F}|\mathcal{A})\mid \mathcal{T}_{m-1},\forall j'\neq 1,\calL_{j'}(f_1)-\calL_{j'}(f)\geq \Delta \calL_{j'}\}}\calL_m(f),
\end{equation}
where event $\mathcal{T}_{m-1}=\{\forall k\in[m-1],\calL_k(f)=\nu_k\}$ and let $\nu_m=\calL_m(f^*)$.

According to the constraints, we have $\calL_1(f^*)=\nu_1$.
Then 
\begin{equation}
\begin{split}
     \epsilon_1(f^*)&=\calL_1(f^*)-\nu_1^0=  \nu_1-\nu_1^0\\
     &=\min_{f\in\{f\in(\mathcal{F}|\mathcal{A})\mid \forall j'\neq 1,\calL_{j'}(f_1)-\calL_{j'}(f)\geq \Delta \calL_{j'}\}}\calL_1(f)-\nu_1^0\\
     &=\min_{f\in\{f\in(\mathcal{F}|\mathcal{A})\mid \forall j'\neq 1,\calL_{j'}(f_1)-\calL_{j'}(f)\geq \Delta \calL_{j'}\}}\epsilon_1(f)=\epsilon_1^0.
\end{split}
\end{equation}

On the other hand, according to the constraints,
$\forall j'\neq 1,\calL_{j'}(f_1)-\calL_{j'}(f^*)\geq \Delta \calL_{j'}$.

Based on the definition of $\nu_1$, if there exists $f'$ such that $\calL_1(f')<\calL_1(f^*)=\nu_1$ and $\calL_{j'}(f')\leq\calL_{j'}(f^*)$ for all $j'\neq 1$, $f'$ should in the space in that $\exists j'\neq 1,\calL_{j'}(f_1)-\calL_{j'}(f')< \Delta \calL_{j'}$,
i.e., $\exists j'\neq 1,\calL_{j'}(f_1)-\Delta \calL_{j'}<  \calL_{j'}(f')$, which, however, generates contradictions with $\calL_{j'}(f')\leq\calL_{j'}(f^*)\leq \calL_{j'}(f_1)-\Delta \calL_{j'} $ for all $j'\neq 1$.

Based on the definition of $\nu_2$, if there exists $f'$ such that $\calL_2(f')<\calL_2(f^*)=\nu_2$ and $\calL_{j'}(f')\leq\calL_{j'}(f^*)=\nu_{j'}$ for all $j'\neq 2$, $f'$ should in the space in that $\calL_1(f')<\nu_1$ or $\exists j'\neq 1,\calL_{j'}(f_1)-\calL_{j'}(f')< \Delta \calL_{j'}$.
By the above proof, $\calL_1(f')<\nu_1$ is not possible.
On the other hand, $\exists j'\neq 1,\calL_{j'}(f_1)-\calL_{j'}(f')< \Delta \calL_{j'}$ means that $\exists j'\neq 1,\calL_{j'}(f_1)- \Delta \calL_{j'}<\calL_{j'}(f')$ which generates a
contradiction with $\calL_{j'}(f')<\calL_{j'}(f^*)\leq \calL_{j'}(f_1)-\Delta \calL_{j'} $ for   $j'= 2$,
and
contradictions with $\calL_{j'}(f')\leq\calL_{j'}(f^*)\leq \calL_{j'}(f_1)-\Delta \calL_{j'} $ for all $j'> 2$.

For other $j'>2$, it can be similarly proved that there does not exist $f'$ such that $\calL_{j'}(f')<\calL_{j'}(f^*)=\nu_{j'}$ and $\calL_{j''}(f')\leq\calL_{j''}(f^*)=\nu_{j''}$ for all $j''\neq j'$.

Therefore, by definition, $f^*$ is on the Pareto front. Due to the symmetry, it holds for each target source $j\in[m]$.
\end{proof}
 
\subsection{Proof of Theorem 2}
\begin{proof}
Without loss of generality, we consider source $1$.
Recall that $\nu_1^0=\min_{f\in\mathcal{F}|\mathcal{A}}\calL_1(f)$. Then constraining on $\calL_1(f)=\nu_1^0$, we find the conditional optimum of $\calL_2$:
\begin{equation}
    \nu_2=\min_{f\in\{f\in(\mathcal{F}|\mathcal{A})\mid \calL_1(f)=\nu_1^0\}}\calL_2(f).
\end{equation}
Continuing this process until source $m$, we get the final model:
\begin{equation}
    f^*=\min_{f\in\{f\in(\mathcal{F}|\mathcal{A})\mid \calL_1(f)=\nu_1^0,\calL_2(f)=\nu_2^0,\ldots,\calL_{m-1}(f)=\nu_{m-1}^0\}}\calL_m(f).
\end{equation}
According to the constraints, we have $\calL_1(f^*)=\nu_1^0$. Then $\epsilon_1(f^*)=\calL_1(f^*)-\nu_1^0=0$. 
On the other hand, $f^*$ cannot decrease any objective further, therefore, is on the Pareto front. Due to the symmetry, it holds for each source.
\end{proof}

\section{Implementation Details}\label{subsec:dt}

\subsection{Details for Our Method}\label{subsec:dt_our}

For the sampling of $\llambda$, we first sample each element in the vector from $\mathcal{U}(0,1)$. Then each element is divided by the sum of all the elements.

For the number of random vectors in the sliced graph matching, in order to approximate the effect of expectation, we sample $128$ random vectors for SWD. The first $8$ vectors are also used for SGWD. Here we only use $8$ vectors, for the sake of efficiency. 
 
For optimizers, we adopt Adagrad. The learning rates are fixed as $0.02$. The hyper-parameter tuning set is $\{10^{-3},10^{-2},10^{-1},1,10\}$ for $\beta$.

For discrete/sparse features, we adopt embedding layers with embedding dim of $8$, whereas, for continuous/dense features, we concatenate them and adopt a fully-connected layer with ELU nonlinear activation as their ``embedding layer''. The output of the embedding layers are fed into a three-layer MLP with hidden dims of $[256,256,32]$ from the bottom to the top. The activations are all ELU. The final layer is $\ell_2$-normalized after activation. We adopt a double-tower structure that the source nodes share an MLP, and the positive/negative nodes share another MLP.
 
For more details, please refer to our publicly available online code at {
\url{https://github.com/anonMLresearcher/ICPA}.}

\subsection{Details for Baseline Methods}\label{subsec:dt_other}

For MMoE, we build an MMoE network for each tower of MLP. MMoE adopts the same number of experts with that of the tasks for fair comparisons. For M2GRL, we also add convolution as in our method.  
The structures of MLPs are the same for MMoE, M2GRL, and Ours. 
For both DGI and GSAGE, we choose a two-layer
structure and dimensions for the first layer and the
second layer are $256$ and $32$, respectively. The batch size is $1024$. The optimizer is Adam.
Other hyper-parameters of each method are tuned according to the strategy mentioned in their respective papers.

\end{document}